\documentclass[conference]{IEEEtran}
\IEEEoverridecommandlockouts
\usepackage{svg}
\usepackage{bbding}
\usepackage{booktabs}
\usepackage{amsmath}
\usepackage{cite}
\usepackage{amsmath,amssymb,amsfonts}

\usepackage{graphicx}
\usepackage{textcomp}
\usepackage{xcolor}
\usepackage{algorithm}
\usepackage{algpseudocode}
\usepackage{multirow} 
\usepackage{makecell}
\def\BibTeX{{\rm B\kern-.05em{\sc i\kern-.025em b}\kern-.08em
    T\kern-.1667em\lower.7ex\hbox{E}\kern-.125emX}}
\begin{document}

\title{Modularized Dynamic-Granularity Video LLM for Multi-Event Long Video Understanding\\
}

\author{ \IEEEauthorblockN{Wei Feng$^{1}$, Xin Wang$^{1,2,*}$, Yu-Wei Zhan$^{1}$, Yuwei Zhou$^{1}$, Wenwu Zhu$^{1,2,*}$}
\IEEEauthorblockA{{$^{1}$Department of Computer Science and Technology, Tsinghua University}\\
$^{2}$Beijing National Research Center for Information Science and Technology, Tsinghua University\\
\{fw22,zhou-yw21\}@mails.tsinghua.edu.cn,zhanyuweilif@gmail.com,\{xin\_wang,wwzhu\}@tsinghua.edu.cn}\thanks{*Corresponding authors: Xin Wang and Wenwu Zhu.}}

\maketitle

\begin{abstract}
Video Large Language Models (Video LLMs) have made significant advancements in various video understanding tasks. However, long-video scenarios remain challenging due to the inherent tension between limited visual token budgets and the need to capture multiple key events. Existing approaches typically process long videos in two stages, i.e., i) select keyframes and ii) perform detailed perception, which exhibit significant limitations: they lack a modular mechanism for adaptive capacity allocation and self-correction, resulting in unreliable modeling. To tackle these challenges, we propose MoD-VLLM, a novel Modularized Dynamic-Granularity Video LLM framework for multi-event long video understanding, which seamlessly unifies temporal grounding and semantic understanding in an iterative, self-reflective manner. Specifically, we propose a Positive-Negative Video Segments Grounding module and a Modularized Dynamic-Granularity Reflection module, which form a closed loop to progressively localize the question-related video segments. The grounding module instructs a Video LLM to distinguish relevant from irrelevant video segments based on the video question. The reflection module employs a modularized scheduler that dynamically selects fine-grained encoding for relevant positive segments to capture detailed perception and coarse-grained encoding for negative segments to efficiently maintain global context, thereby enabling adaptive granularity allocation. We further propose a dynamic-granularity reinforcement learning strategy, allowing MoD-VLLM to learn optimal grounding policies and dynamic granularity visual representation jointly. Moreover, we propose MEventBench, a challenging Multi-Event Long Video Benchmark to evaluate models for complex long video understanding and reasoning. Extensive experiments on several long video understanding benchmarks and our MEventBench demonstrate that the proposed MoD-VLLM is able to significantly outperform state-of-the-art baselines.
\end{abstract}


\section{Introduction}
\label{sec:intro}

Recently, Video Large Language Models (Video LLMs) have made significant advancements. By aligning Large Language Models with vision encoders through instruction tuning, video LLMs are able to perform various video understanding tasks within a unified framework~\cite{liu2024visual,huang2024vtimellm}.

Despite these advances, existing Video LLMs still exhibit fundamental limitations in long video scenarios. Unlike short videos, long videos commonly contain multiple events that are sparsely and discontinuously distributed over time. In essence, multi-event long video understanding presents a fundamental dilemma: the tension between limited visual token budgets and the need to comprehensively capture multiple events. Existing long-video methods fail to resolve this challenge.

\begin{figure}
    \centering
    \includegraphics[width=\linewidth]{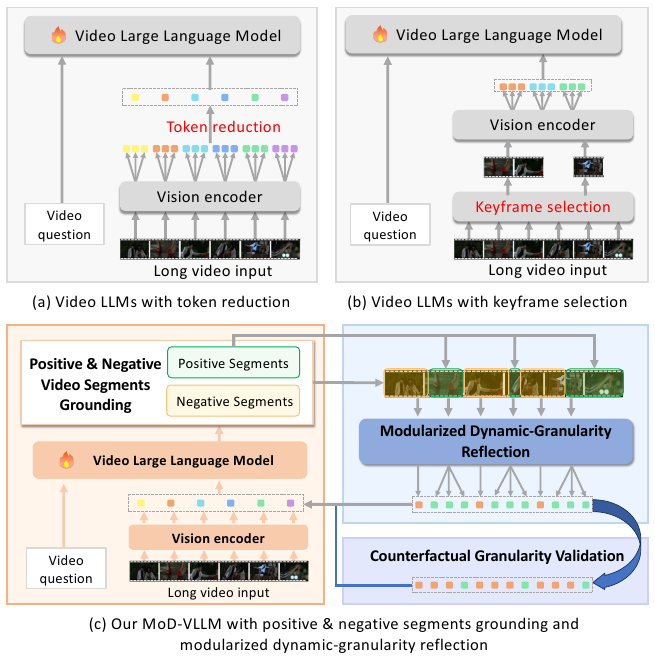}
    \vspace{-0.7cm}
    \caption{Conceptual comparison of different video LLM paradigms. Token reduction methods cause critical detail loss when processing long videos. Keyframe selection suffers from irreversible error propagation when localizing wrong segments. Our MoD-VLLM overcomes these issues via iterative, self-corrective grounding and modularized encoding with dynamic granularity.}
    \vspace{-0.7cm}
    \label{fig:concept}
\end{figure}

Specifically, approaches shown in Figure~\ref{fig:concept}(a) employ token pruning or frame sampling strategies to reduce the visual tokens~\cite{song2024moviechat,shenlongvu}. Such static visual modeling strategies overlook the semantic importance differences among segments and fail to capture fine-grained details that are critical for answering the query.
Other methods shown in Figure~\ref{fig:concept}(b) adopt two-stage strategies~\cite{tang2025adaptive,wang2025videotree}, where keyframes are first localized and then processed for detailed understanding. These approaches lack effective self-reflection and error-correction during localization, causing errors to accumulate once initial predictions are biased.
As a result, existing methods lack a mechanism to adaptively allocate modeling capacity based on segment importance and fail to perform effective self-correction, leading to unreliable modeling of multiple key events.

To address these challenges, as shown in Figure~\ref{fig:concept}(c), we propose \textbf{MoD-VLLM}, a novel \textbf{Mo}dularized \textbf{D}ynamic-Granularity \textbf{V}ideo \textbf{LLM} for multi-event long video understanding. 
Specifically, our MoD-VLLM framework consists of i) a {Positive-Negative Video Segments Grounding} module, which first instructs the video LLM to identify and distinguish relevant and irrelevant video segments regarding the multi-event question, and ii) a {Modularized Dynamic-Granularity Reflection} module, which adopts a modular scheduling mechanism to adaptively allocate dynamic granularity across video frames. It performs fine-grained encoding on relevant segments to capture detailed semantics, while applying coarse-grained encoding to irrelevant segments to efficiently maintain global visual context.
The reflection module provides a closed-loop feedback mechanism, where counterfactual granularity verification serves as a basic reflection process to mitigate localization errors, progressively guiding the model toward more accurate video segment localization.
To optimize the proposed MoD-VLLM framework, we further propose a dynamic-granularity reinforcement learning strategy for optimal grounding policies generation and dynamic granularity representation learning. In addition, we construct and propose \textbf{MEventBench}, a \textbf{M}ulti-\textbf{Event} \textbf{Bench}mark for long video understanding, which contains long videos with questions about several key video segments and temporal dependencies. Extensive experiments on several long video understanding benchmarks and our MEventBench show that the proposed MoD-VLLM framework is able to significantly outperform existing state-of-the-art baselines.

In summary, we make the following contributions:
\begin{itemize}
    \item We propose MoD-VLLM, a modularized dynamic-granularity video LLM framework for accurate multi-event long video understanding, which is capable of conducting dynamic granularity iteration through positive-negative video segments grounding and modularized dynamic-granularity reflection.
    \item We propose a dynamic-granularity reinforcement learning strategy to optimize the MoD-VLLM framework.
    \item We propose MEventBench, a multi-event benchmark for long video understanding. 
    \item We conduct extensive experiments on several long video understanding benchmarks and our MEventBench, which demonstrate that MoD-VLLM can outperform state-of-the-art baseline methods.
\end{itemize}

\begin{figure*}[h]
    \centering
    \includegraphics[width=\linewidth]{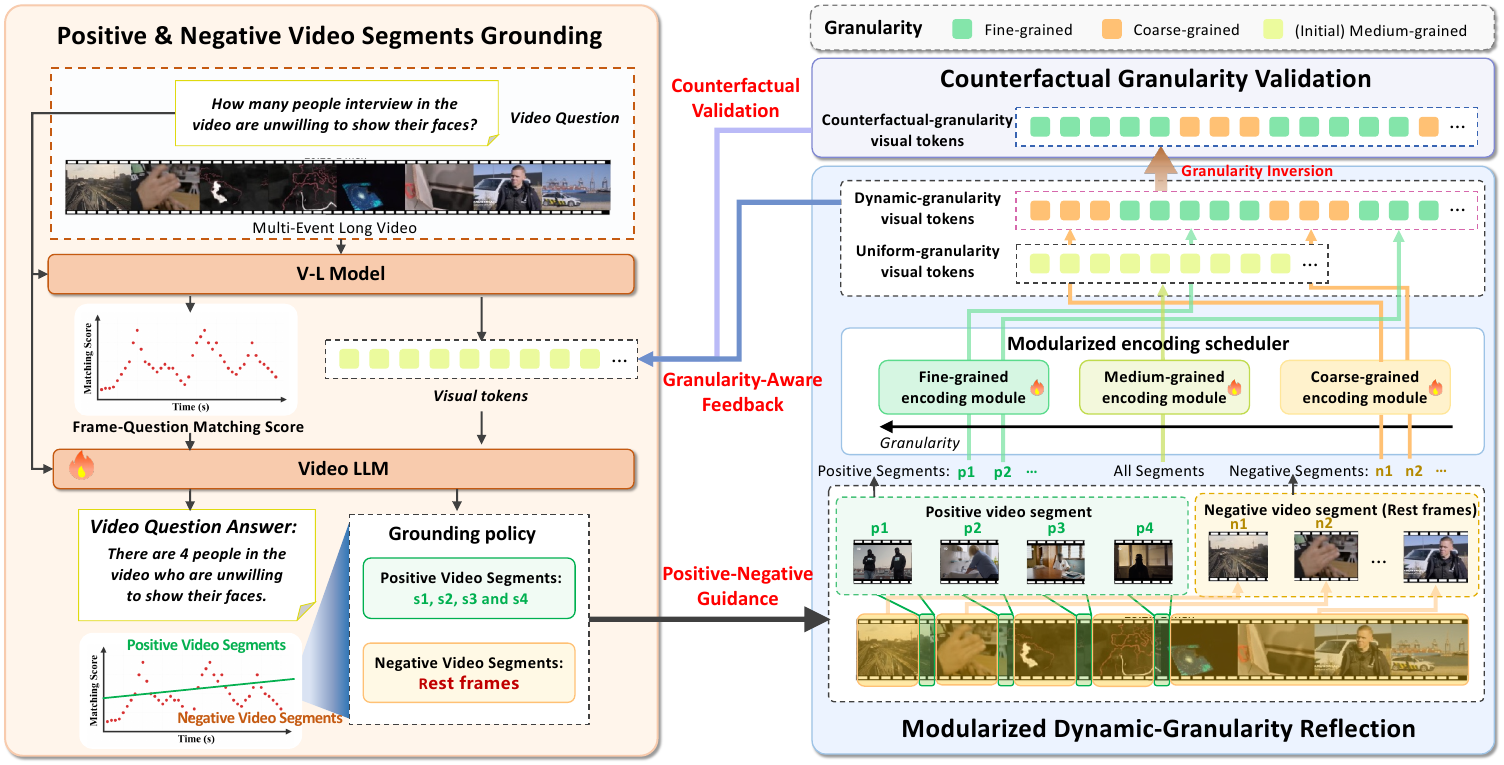}
    \vspace{-0.8cm}
    \caption{Our MoD-VLLM framework for multi‑event long video understanding. The positive‑negative video segments grounding module instructs the video LLM to generate a grounding policy that identifies question‑relevant segments. The modularized dynamic‑granularity reflection module then differentiates encoding granularity: positive segments are encoded in a fine‑grained manner for detail, while negative segments are encoded coarsely to retain global context. Through iterative updates of the visual token sequence, the model refines grounding policies and answers. An optional counterfactual granularity sequence (with reversed granularity assignments) is used for validation to mitigate error propagation.}
    \vspace{-0.7cm}
    \label{fig:framework}
\end{figure*}
\section{Related Work}
\label{sec:related}

Recently, advances in Large Language Models~\cite{touvron2023llama} have led to the development of Video LLMs for temporal video understanding through cross-modal alignment~\cite{wang2025multi}. Models such as Video-LLaMA~\cite{zhang2023video} typically employ visual encoders to extract frame-level features and project them into the LLM's space, sharing ideas with image-based LLMs~\cite{bavishi2023fuyu}.
However, when dealing with long videos, these models face a critical challenge: processing all frames at high sampling rates exceeds token limits, while reducing sampling rates risks losing essential temporal information and visual details.

To address the token limitation in long video understanding, some methods focus on compressing visual tokens. Video LLMs like LongVU~\cite{shenlongvu} reduce token counts, while MovieChat~\cite{song2024moviechat} uses memory mechanisms for efficient compression. 
However, these approaches often discard fine-grained spatial-temporal details needed for complex video reasoning.
Other methods adopt a two-stage, coarse-to-fine strategy. Approaches like VideoTree~\cite{wang2025videotree} and Adaptive Keyframe Selection(AKS)~\cite{tang2025adaptive} first identify key segments coarsely, then analyze them in detail. While this strategy suffers from error propagation if initial localization fails, and the two stages are typically trained separately, leading to suboptimal alignment between localization and understanding.

\section{Method}
We introduce our MoD-VLLM framework in this section.
Inspired by the application of video grounding for different video tasks~\cite{wang2025video,feng2025multi}  and modularization designs for the multimodal large language models~\cite{zhan2025bitagent}, we introduce similar approaches to advance multi‑event understanding in long videos.
As shown in Figure~\ref{fig:framework}, MoD-VLLM designs a Positive-Negative Video Segments Grounding module and a Modularized Dynamic-Granularity Reflection module, which operate iteratively for multi-event long video understanding. Given a long video input, the grounding module first instructs the video LLM to identify video segments relevant to the question (positive segments) and treat the remaining as irrelevant (negative segments). Based on this guidance, the reflection module then employs a modularized encoding scheduler that dynamically selects from a set of pre‑defined granularity modules (varying in per‑frame token counts and sampling rates). It represents positive segments with fine‑grained tokens for detailed perception, while encoding negative segments coarsely to preserve global context. This dynamic granularity representation replaces the original uniform‑granularity tokens and is fed back to the grounding module for iterative refinement. Over several iterations, the video LLM progressively refines both the grounding policy and the final answer to the video question.

\subsection{Positive-Negative Video Segments Grounding}

Our grounding module uses two branches: uni‑granularity representation and question‑frame similarity. The video LLM then identifies question‑relevant segments as positive and the rest as negative.

\textbf{Uni‑granularity representation.} Given a long video $v\in R^{T\times H\times W\times C}$, we uniformly sample $N$ frames $\widetilde{v}\in R^{N\times H\times W\times C}$. Each frame is encoded by a vision transformer:
\begin{equation}
    \{v_i^{cls},v_i^{1},v_i^{2},\dots,v_i^{patch}\}=ViT(\widetilde{v}_i),\quad i=1,\dots,N,
\end{equation}
and the global feature $v_i^{cls}$ is projected into visual tokens:
\begin{equation}
\begin{split}
    z_i &= f(v_i^{cls}),\quad i=1,\dots,N,\\
    Z   &= \{z_i\} \in R^{N\times L\times d},
\end{split}
\end{equation}
where $Z$ is the visual token sequence, $d$ is the hidden dimension, and $L$ is the tokens per frame.

\textbf{Question‑frame similarity representation.} Using the same frames and features, we compute similarity between $v_i^{cls}$ and question embedding $q$:
\begin{equation}
\begin{split}
    sim_i &= \frac{v_i^{cls}}{\|v_i^{cls}\|} \frac{q}{\|q\|},\\
    Sim   &= \{sim_i\} \in R^{N}.
\end{split}
\end{equation}

\textbf{Multi-event video grounding.} To combine the two obtained branches' visual information into the video LLM, we design a structured input paradigm that systematically integrates visual tokens with the question-frame similarity guidance. Specifically, we formulate this through a multi-modal prompt template: 
\begin{equation}
\begin{split}
\label{eq:input1}
    input &= Concat(p_{inst},Z,p_{Sim},p_{task}),
\end{split}
\end{equation}
where $p_{inst}$ denotes the instruction prompt `Based on the following visual tokens and frame-query similarity scores:', $p_{Sim}$ denotes the textual embeddings of the above question-frame similarity sequence, and $p_{task}$ denotes the task prompt `Identify which parts of the video segments are relevant to the question query $q$'. The complete prompt also includes some other examples of outputs and the necessary description for the input video information, such as video duration. This design enables the video LLM to simultaneously process visual semantics and similarity guidance. The question-frame similarities with higher scores would directly guide the video LLM to pay attention to the fact that these video segments are semantically close to the long video question.
The final segments' grounding output would be responded to by the video LLM in a restricted JSON format:
\begin{equation}
\begin{split}
\label{eq:grounding_output}
    output&=VidLLM(input)\\
    &=[[s_1,e_1],...,[s_j,e_j]],
\end{split}
\end{equation}
where $s$ and $e$ denote the start time and end time for one grounding segment, and the number of $j$ is irregular depending on the related video events to the question. These video segments are currently considered positive video segments more related to the video question, while the rest are considered negative video segments.
\subsection{Modularized Dynamic-Granularity Reflection}

Based on the grounding results, we apply dynamic granularity encoding to differentiate positive and negative segments.

\textbf{Modularized encoding modules with dynamic granularity.} 
We build a set of encoding modules $\mathcal{E} = \{\phi_1, \dots,\phi_{b-1},\phi_b,\phi_{b+1},\dots, \phi_K\}$ by fine-tuning the projection layers between the visual encoder and the LLM. A shared ViT backbone extracts features, while switchable projections produce tokens at different granularities. By combining modules with different per‑frame token budgets and sampling rates, we can flexibly tailor the information density of the visual representation to meet diverse video understanding requirements. $\phi_m = \phi_b$ represents the initial uniform encoding granularity on the video segments grounding. Coarser module $\phi_c=\phi_{b-1}$ reduces tokens per frame for compression and finer modules $\phi_f=\phi_{b+1}$ increases tokens for detail. 
$\{\phi_i, i<b-1\}$ denotes the coarse-grained encoding module applying a lower frame sampling rate compared to the initial setting, while $\{\phi_i, i>b+1\}$ denotes the fine-grained encoding module applying a higher frame sampling rate, and their sampling rates are sorted from low to high. Based on the constructed group of encoding modules, we can assert that if $p>n$, then encoding module $\phi_p$ would contain richer visual information than $\phi_n$. 

\textbf{Modularized encoding scheduler.} For detailed perception on the key video segments, we aim to encode positive segments as fine‑grained as possible without exceeding the LLM's token limit $L_{max}$. Let $r_k$ be the token generation rate (tokens/frame) of module $\phi_k$. After grounding, we obtain positive frame count $T_p$ and negative frame count $T_n$, with proportion $\rho = T_p / T$. Granularity levels for negative and positive segments are computed as:
\begin{equation}
\begin{split}
    \text{GranularityLevel}_{neg}(\rho) &= b - \left\lfloor (b-1) \cdot \rho^\alpha \right\rfloor,\\
    \text{GranularityLevel}_{pos}(\rho) &= b + \left\lfloor (K-b) \cdot (1-\rho^\alpha) \right\rfloor,
\end{split}
\end{equation}
where $\alpha\in[1.5,2.0]$ is an adjustable factor. The corresponding encoding modules are selected as:
\begin{equation}
\begin{split}
    \phi_n &= \phi_{\max(1,\min(\text{GranularityLevel}_{neg}(\rho),b))},\\
    \phi_p &= \phi_{\min(K,\max(\text{GranularityLevel}_{pos}(\rho),b))}.
\end{split}
\end{equation}
The total token budget must satisfy:
\begin{equation}
    T_p \cdot r_p + T_n \cdot r_n \leq L_{max}.
\end{equation}
If the budget is exceeded, we first downgrade $\phi_n$ (negative segments) to a coarser module, then $\phi_p$ if needed. Conversely, if tokens are far below $L_{max}$, we upgrade $\phi_p$ first. This can be summarized as the optimization:
\begin{equation}
\begin{aligned}
& \underset{\phi_p, \phi_n \in \mathcal{E}}{\text{maximize}} 
& & w_p \cdot r_p - w_n \cdot r_n, \\
& \text{subject to}
& & T_p \cdot r_p + T_n \cdot r_n \leq L_{\text{max}}, \\
& & & r_p \geq r(\phi_{\text{base}}) \geq r_n, \\
& & & \phi_p \in \{\phi_b, \phi_{b+1}, \dots, \phi_K\}, \\
& & & \phi_n \in \{\phi_1, \phi_2, \dots, \phi_b\},
\end{aligned}
\end{equation}
where $w_p$, $w_n$ are preference weights for positive/negative segments.

Given the selected modules, we encode each continuous video segment $s_i$ with its assigned module $\phi_i \in \{\phi_p, \phi_n\}$, producing the final dynamic granularity token sequence:
\begin{equation}
    Z' = \{\phi_1(s_1), \phi_2(s_2), \dots, \phi_t(s_t)\},
\end{equation}
which replaces the original uniform token sequence $Z$ and is fed back to the grounding module for iterative refinement.

\begin{table*}[ht]

    \centering
    \caption{Performance comparison on long video understanding benchmarks. * denotes applying GPT-4 as LLM for the main result. 
    }
    \vspace{-0.3cm}
    \begin{tabular}{lcccccccc}
    \toprule
        \multirow{2}{*}{Models}  &\multirow{2}{*}{Size}  &\multicolumn{4}{c}{VideoMME}&\multirow{2}{*}{Lvbench}&\multirow{2}{*}{MLVU Dev} &\multirow{2}{*}{MEventBench}\\ \cmidrule(lr){3-6} && Short&Medium&Long&Overall \\
    \midrule
    Duration(min)&& $\leq$2&4$\sim$15&30$\sim$60&1$\sim$60&30$\sim$140&3$\sim$120&15$\sim$100\\
    \midrule    
    
    Video-RAG~\cite{luo2024video}&7B&66.4&60.2& 59.8& 62.1&39.3&65.2&52.5\\
    LongVU~\cite{shenlongvu}&7B&64.7 &{58.2}& {59.5}& {60.9}&{38.3}&{65.4}&51.2\\
    Video-XL~\cite{shu2025video}&7B&{67.4}&{60.7}&54.9&{61.0}&37.7&64.9&49.3\\
    Video-XL2~\cite{qin2025video}&8B&{73.7}&{65.9}&60.2&{66.6}&\underline{48.4}&74.8&53.8\\
    VITA 1.5~\cite{fu2025vita}&7B&67.0&54.2&47.1&56.1&32.1&60.2&45.9\\
    AKS~\cite{tang2025adaptive}&7B&-&-&-&65.4&46.8&70.1&55.7\\
    VideoTree~\cite{wang2025videotree}&*&-&-&54.2&-&-&-&55.4\\
    Qwen2.5-VL~\cite{bai2025qwen2}&7B&\textbf{81.4}&\underline{70.8}&\underline{62.6}&\underline{71.6}&45.3&\underline{75.5}&\underline{60.8}\\
    \midrule
    LLaVA-Video~\cite{zhang2024video}&7B&78.0&69.3&61.8&69.7&43.2&71.4&59.2\\
    Ours-MoD-VLLM w/ LLaVA-Video&7B &\underline{80.3}&\textbf{72.4}&\textbf{66.9}&\textbf{73.2}&\textbf{49.6}&\textbf{78.2}&\textbf{64.8}\\
    \bottomrule
    \vspace{-1.0cm}
    \end{tabular}
    \label{tab:video_mme}
\end{table*}

\subsection{Dynamic Granularity Iteration with RL}

We combine the grounding and reflection modules into an iterative framework, and optimize the MoD-VLLM via a dynamic granularity reinforcement learning strategy.

\textbf{Dynamic granularity iteration.} To enable multi‑event understanding, we iteratively alternate between grounding and reflection. After the first iteration, we replace the original uniform token sequence $Z$ with the dynamic granularity tokens $Z'$ for subsequent grounding. Two multimodal inputs are constructed:
\begin{equation}
\begin{split}
    input_{grounding} &= Concat(p_{inst}, Z', p_{Sim}, p_{task}),\\
    input_{answering} &= Concat(Z', q),
\end{split}
\end{equation}
where $input_{grounding}$ follows the same structure as Eq.~\ref{eq:input1}, and $q$ is the original question. Both inputs include textual instructions about the temporal arrangement of the interleaved tokens. The answer is obtained by feeding $input_{answering}$ to the video LLM.

To mitigate error propagation from missed relevant segments (which would be encoded coarsely), we introduce a counterfactual validation in the first reflection step. In addition to $Z'$, we generate a counterfactual token sequence:
\begin{equation}
    Z'' = \{\phi_1'(s_1), \phi_2'(s_2), \dots, \phi_t'(s_t)\},
\end{equation}
where $\phi_i'$ is assigned by the scheduler under reversed positive‑negative classification. $Z''$ is used only in the first iteration. During the second grounding step, the model processes both $Z'$ and $Z''$ and merges the two grounding outputs to reduce error propagation.

\textbf{Dynamic granularity reinforcement learning.} To improve grounding accuracy, we optimize our MoD-VLLM framework using direct preference optimization (DPO) within our iterative framework. During grounding, the model is instructed to generate multiple candidate policies $\{p_i\}$ (each in the JSON format of Eq.~\ref{eq:grounding_output}). Each policy leads to a dynamic token sequence $Z_i$ via the scheduler.

A high‑quality token sequence should represent question‑relevant frames in detail while compressing irrelevant parts. Therefore, feeding a well‑structured $Z_i$ to the video LLM should produce an answer closer to the ground truth, having a lower cross‑entropy loss. Conversely, a poor sequence would increase loss.
Inspired by RLHF~\cite{christiano2017deep},
we apply DPO~\cite{rafailov2024direct,song2025modularized} using the cross‑entropy scores as implicit preferences. Let $p_w$ be the policy with the smallest loss and $p_l$ a policy with a larger loss. The DPO objective is:
\begin{equation}
\label{equ:dpo}
\small
\begin{split}
    &L_{DPO}(\pi_{\theta};\pi_{ref})=
    \\&-E_{(q,v,p_w,p_l)\sim \mathcal{D}}[\log \sigma(\beta\log\frac{\pi_{\theta}(p_w|q,v)}{\pi_{ref}(p_w|q,v)}-\beta\log\frac{\pi_{\theta}(p_l|q,v)}{\pi_{ref}(p_l|q,v)})],
\end{split}
\end{equation}
where $\pi_\theta$ is the trainable video LLM, $\pi_{ref}$ is a fixed reference video LLM, $\sigma$ is the sigmoid function, $\beta$ is a scaling parameter, and $\mathcal{D}$ is the training dataset.

\vspace{-0.2cm}
\section{Experiment}

\textbf{Implementations.} We use LLaVA-Video(7B)~\cite{zhang2024video} as the video LLM backbone. Fine‑tuned encoding modules $\phi_{coarse}$, $\phi_{medium}$, $\phi_{fine}$ produce 36, 64, and 169 tokens per frame respectively. For initial grounding, we compute similarity scores with CLIP~\cite{radford2021learning} and sample 128 frames using the medium‑grained encoder (64 tokens/frame), with initial sampling rate $N_T = 128 / T_{video}$. The full encoding set includes \{$(0.2N_T,\phi_{coarse})$, $(0.5N_T,\phi_{coarse})$, $(N_T,\phi_{coarse})$, $(N_T,\phi_{medium})$, $(N_T,\phi_{fine})$, $(2N_T,\phi_{fine})$, $(3N_T,\phi_{fine})$\}. We run three dynamic granularity iterations per video. Further parameters and training details are in the supplement.

\textbf{Evaluation.} We evaluate on long‑video benchmarks: VideoMME~\cite{fu2024video}, Lvbench~\cite{wang2024lvbench}, and MLVU~\cite{zhou2025mlvu}, reporting accuracy on multiple‑choice questions. To assess complex multi‑event reasoning, we construct \textbf{MEventBench}, containing 1200 video‑question pairs from VideoMME, Longvideobench~\cite{wu2024longvideobench}, InfiniBench~\cite{ataallah2024infinibench}, and CG‑Bench~\cite{chen2025cg}. Each video includes at least three question‑related segments dispersed in time. 
In summary, we classify the type of our selected multi-event data into multi-event counting, ordering, and reasoning. More details are provided in the supplement.

\subsection{Main Result on Long Video Understanding}

Table~\ref{tab:video_mme} reports results on long‑video benchmarks. Our MoD-VLLM achieves higher average accuracy on longer videos (>2min), outperforming state‑of‑the‑art methods of similar scale. These include open‑source video LLMs (LongVU~\cite{shenlongvu}, Video‑XL2~\cite{qin2025video}, Qwen2.5‑VL~\cite{bai2025qwen2}) and two‑stage coarse‑to‑fine approaches (Video‑RAG~\cite{luo2024video}, AKS~\cite{tang2025adaptive}, VideoTree~\cite{wang2025videotree}). On our MEventBench, MoD-VLLM also surpasses the strongest baseline, validating the effectiveness of dynamic granularity iteration for complex multi‑event reasoning.


\vspace{-0.1cm}
\subsection{Ablation Study}
In this section, we provide ablation results about our concerned questions as follows.

\begin{table}[h]
    \centering
    \vspace{-0.4cm}
    \caption{Ablation study on different multi-event tasks.}
    \vspace{-0.3cm}
    \begin{tabular}{ccccc}
    \toprule
        \multirow{2}{*}{Method} &  \multicolumn{4}{c}{MEventBench}\\
        \cmidrule(lr){2-5}&Counting&Ordering&Reasoning&Overall\\
        \midrule
        Qwen2.5-VL&64.2&61.1&57.2&60.8\\
        LLaVA-Video&62.0&59.4&56.3&59.2\\
        MoD-VLLM&{69.4}&64.2&62.1&64.8\\    
         \bottomrule
    \end{tabular}
    \vspace{-0.3cm}
    \label{tab:meventbench}
\end{table}
\textbf{Performance on different multi‑event tasks.} Table~\ref{tab:meventbench} shows results across task categories on MEventBench. MoD-VLLM outperforms strong video LLM baselines in all categories, achieving an overall accuracy of 64.8\%, which is 4.0\% and 5.6\% higher than Qwen2.5‑VL and LLaVA‑Video, respectively. The largest gain appears in Counting (69.4\%, +5.2\% over Qwen2.5‑VL), highlighting our dynamic granularity iteration's ability to localize sparse, distributed events. Improvements in Ordering (64.2\%) and Reasoning (62.1\%) further confirm the framework's strength in modeling temporal relations and complex inference, enabled by iterative refinement and positive‑negative context modeling. These results demonstrate that allocating fine‑grained representation to critical segments while maintaining global context is key to comprehensive long‑video understanding.

\begin{table}[h]
    \centering
    \scriptsize
    \vspace{-0.5cm}
    \caption{Ablation study on input modalities for grounding.}
    \vspace{-0.3cm}
    \begin{tabular}{ccccc}
    \toprule
    \multirow{2}{*}{Method} &\multirow{2}{*}{VideoMME} & \multicolumn{3}{c}{MEventBench}\\
        \cmidrule(lr){3-5}&&Counting&Ordering&Reasoning\\
        \midrule
         MoD-VLLM w/ V+S& 73.2&69.4&64.2&62.1\\
         MoD-VLLM w/ V&67.1&65.8&60.3&57.7\\
         MoD-VLLM w/ S&62.5&59.6&55.0&51.8\\
         \bottomrule
    \end{tabular}
    \vspace{-0.4cm}
    \label{tab:modality}
\end{table}
\textbf{Why apply both visual representation and similarity sequence for grounding instruction?} 
We evaluate the contribution of visual tokens (V) and similarity scores (S) to grounding. As Table~\ref{tab:modality} shows, using both (V+S) achieves the best performance across VideoMME and all MEventBench tasks. Relying only on visual tokens (V) or only similarity (S) leads to clear drops, with similarity‑alone performing weakest.

The advantage of V+S stems from two factors. First, CLIP‑based similarity, trained on short image‑text pairs, does not fully capture the alignment between long questions and video frames. Second, without similarity as a prior, the video LLM initially trained for QA, struggles to ground multiple segments under DPO, due to the cold start in a more complex task. The visual tokens provide dense semantics while the similarity scores offer a direct question‑frame relevance signal, enabling robust mutual verification and improving grounding accuracy.

\begin{figure}[h]
    \centering
    \vspace{-0.4cm}
    \includegraphics[width=0.95\linewidth]{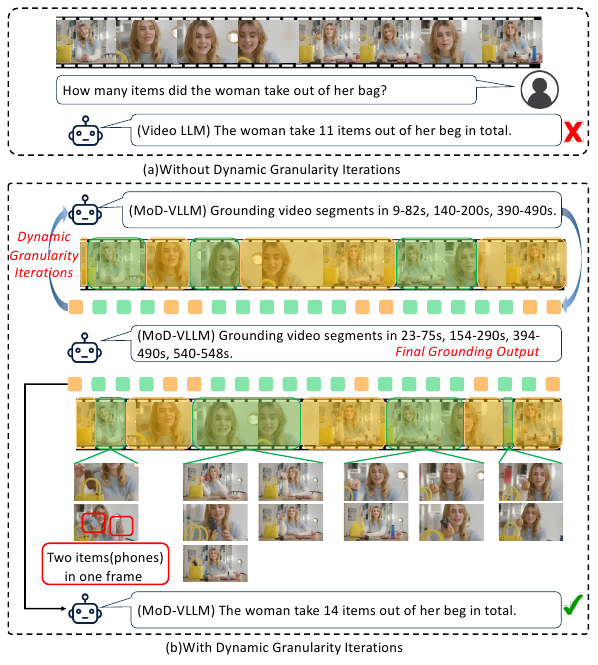}
    \vspace{-0.5cm}
    \caption{A qualitative example of multi-event long video understanding. The duration is about 10 minutes long.}
    \vspace{-0.6cm}
    \label{fig:case}
\end{figure}

\textbf{Case Analysis.}
To demonstrate MoD-VLLM's complex video reasoning, Figure~\ref{fig:case} visualizes a multi‑event counting example with the question: ``How many items did the woman take out of her bag?" The task is challenging because the key segments are distributed over time and require detailed perception (e.g., two phones are taken simultaneously in one segment). Through dynamic granularity iteration, our model is able to progressively refine its attention and produce a granularity encoding that highlights all relevant segments where items are removed.

\section{Conclusion}
In this paper, we propose MoD-VLLM, a novel modularized dynamic-granularity video LLM framework with reflection for multi-event long video understanding. Specifically, we design the Positive-Negative Video Segments Grounding module and the Modularized Dynamic-Granularity Reflection module to enable dynamic granularity iteration on complex long videos. We further propose a dynamic-granularity reinforcement learning strategy to optimize MoD-VLLM with dynamic granularity representation. We construct and propose MEventBench, a multi-event benchmark for long video understanding. Extensive experiments on several long video understanding benchmarks and our MEventBench demonstrate that the proposed MoD-VLLM framework outperforms existing state-of-the-art baselines on processing complex long videos with multiple question-relevant segments.

\vspace{-0.3cm}
\bibliographystyle{IEEEtran}
\bibliography{ref}

\newpage
\section*{Supplement}
\subsection{Training details}
The overall training of our framework is conducted under 8 NVIDIA A100-40GB GPUs.
Our framework is built upon the LLaVA-Video architecture as the backbone model. To achieve dynamic granularity control, we introduce additional adaptive pooling modules in the projection layers between the visual encoder and the LLM, which systematically regulate the number of tokens per video frame. This allows us to construct three distinct encoding modules operating at different granularity levels: 36 tokens per frame (coarse), 64 tokens per frame (medium), and 169 tokens per frame (fine), representing progressively increasing spatial detail. For the dynamic granularity encoding scheduler, we set the adjustable factor $\alpha$ to 1.5.

\textbf{Module training details.}
For training the encoding modules in the first stage, we employ the LLaVA-Video-178K~\cite{zhang2024video} dataset to finetune the video LLM. During the reinforcement learning stage, we employ the VideoITG dataset~\cite{wang2025videoitg}, which contains 40K videos and 500K instruction-guided annotations.

\begin{table}[h]
    \centering
    \scriptsize
    \caption{Implementation details}
    \begin{tabular}{ccccccccccccc}
    \toprule
    Config&\multicolumn{3}{c}{Finetuning}&Reinforcement Learning\\
    \midrule
    Video LLM & \multicolumn{4}{c}{LLaVA-Video-7B}\\
    Optimizer&\multicolumn{4}{c}{AdamW}\\
    Epochs&\multicolumn{3}{c}{10}&5\\
    Warmup ratio&\multicolumn{4}{c}{0.05}\\
    Learning rate	&\multicolumn{3}{c}{2e-5}&1e-5\\
    Batch size&\multicolumn{4}{c}{1}\\
    Gradient accumulation steps	&\multicolumn{4}{c}{16}\\
    Number of Tokens per Frame &36&64&169&mixed\\
    Training cost	&\multicolumn{3}{c}{380h in total}&{800h}\\
    \bottomrule
    \end{tabular}
    \label{tab:implementation}
\end{table}

\subsubsection{Algorithm details}
\textbf{Modularized encoding scheduler.} 
Compared to the original uni-granularity representation, we hope to represent the positive video segment as fine-grained as possible with a higher ratio of sampled frames, but not exceeding the token limitations $L_{max}$\footnote{In actual implementation, $L_{max}$ would be smaller than the max token limitations of LLM, to preserve part of space for processing long text prompt.} of the video LLM. Defining $r_k$ is the token generation rate with the number of frames (tokens/frame) for the encoding module $\phi_k$, since we are aware that the initial video segments grounding is executable with the medium-grained encoding module, so we have:
\begin{equation}
\begin{split}
    T\cdot r_{base}\lesssim L_{max},
\end{split}
\end{equation}
which represents the number of visual tokens is less than, but very close to the limit of LLM. Based on the grounding guidance, we can acquire the total number of positive video frames $T_p$ and negative video frames $T_n$. Considering $\rho=\frac{T_p}{T}$ as the proportion of positive sample frames, we explore the granularity level as:
\begin{equation}
\begin{split}
    \text{GranularityLevel}_{neg}(\rho) &= b - \left\lfloor (b-1) \cdot \rho^\alpha \right\rfloor,\\
    \text{GranularityLevel}_{pos}(\rho) &= b + \left\lfloor (K-b) \cdot (1-\rho^\alpha) \right\rfloor,
\end{split}
\end{equation}
where $\alpha$ is an adjustable factor ranging from 1.5 to 2.0. According to the granularity level, we select the encoding module for positive segments and negative segments separately:
\begin{equation}
\begin{split}
    \phi_n&=\phi_{max(1,min(\text{GranularityLevel}_{neg}(\rho),b))},\\
    \phi_p&=\phi_{min(K,max(\text{GranularityLevel}_{pos}(\rho),b))}.
\end{split}
\end{equation}
Since the total token limitations should be fulfilled as:
\begin{equation}
\begin{split}
    T_p\cdot r_p+T_n\cdot r_n\leq L_{max}.
\end{split}
\end{equation}
And we design our scheduler algorithm as shown in  Algorithm~\ref{alg:granularity_scheduler}. When we notice the total number of required tokens $L_{total}$ higher than $L_{max}$, we would first try to downgrade the encoding module for negative segments to a more coarse-grained one, and if necessary downgrade the encoding module for positive segments, utill $L_{total}\leq L_{max}$. Meanwhile, if the $L_{total}$ is far less than $L_{max}$, we will try to upgrade the granularity for the positive encoding module first.

\begin{algorithm}
\caption{Dynamic Granularity Encoding Scheduler}
\label{alg:granularity_scheduler}
\begin{algorithmic}[1]
\Require $T_p, T_n, \mathcal{E}, L_{\text{max}}, \phi_p, \phi_n, \phi_{\text{base}}=\phi_b=\phi_{medium}$
\Ensure $\phi_p^*, \phi_n^*$

\State $\phi_p^* \gets \phi_p, \phi_n^* \gets \phi_n$
\State $L \gets T_p \cdot r(\phi_p^*) + T_n \cdot r(\phi_n^*)$

\While{$L > L_{\text{max}}$}
    \If{$\phi_n^* \neq \phi_1$}
        \State $\phi_n^* \gets \text{coarser}(\phi_n^*)$
    \ElsIf{$\phi_p^* \neq \phi_{\text{base}}$}
        \State $\phi_p^* \gets \text{coarser}(\phi_p^*)$
    \Else
        \State \textbf{break}
    \EndIf
    \State $L \gets T_p \cdot r(\phi_p^*) + T_n \cdot r(\phi_n^*)$
\EndWhile

\While{$L \ll L_{\text{max}}$}
    \State $\phi_p^{\text{temp}} \gets \phi_p^*, \phi_n^{\text{temp}} \gets \phi_n^*$
    
    \If{$\phi_p^{\text{temp}} \neq \phi_K$}
        \State $\phi_p^{\text{temp}} \gets \text{finer}(\phi_p^{\text{temp}})$
    \ElsIf{$\phi_n^{\text{temp}} \neq \phi_{base}$}
        \State $\phi_n^{\text{temp}} \gets \text{finer}(\phi_n^{\text{temp}})$
    \Else
        \State \textbf{break}
    \EndIf
    
    \State $L_{\text{new}} \gets T_p \cdot r(\phi_p^{\text{temp}}) + T_n \cdot r(\phi_n^{\text{temp}})$
    \If{$L_{\text{new}} \leq L_{\text{max}}$}
        \State $\phi_p^* \gets \phi_p^{\text{temp}}, \phi_n^* \gets \phi_n^{\text{temp}}, L \gets L_{\text{new}}$
    \Else
        \State \textbf{break}
    \EndIf
\EndWhile

\State \Return $\phi_p^*, \phi_n^*$
\end{algorithmic}
\end{algorithm}

Overall, the selection based on the ratio of positive segments could be summarized as the following optimization problem:
\begin{equation}
\begin{aligned}
& \underset{\phi_p, \phi_n \in \mathcal{E}}{\text{maximize}} 
& & w_p \cdot r_p - w_n \cdot r_n, \\
& \text{subject to}
& & T_p \cdot r_p + T_n \cdot  r_n \leq L_{\text{max}}, \\
& & & r_p \geq r(\phi_{\text{base}}) \geq r_n ,\\
& & & \forall r_k = r(\phi_k),\phi_{base}=\phi_b,\\
& & & \phi_p \in \{\phi_b, \phi_{b+1}, \dots, \phi_K\} ,\\
& & & \phi_n \in \{\phi_1, \phi_2, \dots, \phi_b\},
\end{aligned}
\end{equation}
where $w_p$ and $w_n$ are the weight coefficients representing the preference for the positive video segments.

Obtaining the suitable encoding modules for positive video segments and negative ones, our dynamic granularity encoding would process video frames through interleaved fine-grained and coarse-grained representations according to the temporal distribution of positive and negative video segments. Given the partitioned and continus video segments $\{s_1,s_2,...,s_t\}$ with corresponding granularity assignments $\{\phi_1,\phi_2,...,\phi_t\}$ where $\phi_i\in\{\phi_p,\phi_n\}$, we form the final dynamic encoding visual tokens:
\begin{equation}
\begin{split} 
    Z'&=\{\phi_1(s_1),\phi_2(s_2),...,\phi_t(s_t)\},
\end{split}
\end{equation}
where $Z'$ would serve as granularity-aware feedback reflected to the positive-negative video segments grounding module, replacing the original input visual sequence $Z$.

\textbf{Dynamic granularity reinforcement learning.} Since the applied video LLM is not trained with localizing multiple related events in long videos, it is difficult to ensure the accuracy of the grounding policy through in-context prompt learning alone. To address that, we design a dynamic granularity reinforcement learning method to optimize the MoD-VLLM framework through our dynamic granularity iteration framework.

To begin with, during the positive-negative video segments grounding strategy, we prompt the video LLM to generate several multi-event grounding policies $\{p_i\}$, each policy $p_i$ is in the same JSON format as $output$ in equation~\ref{eq:grounding_output}. Each policy corresponds to a group of positive video segments and negative ones, which would inform the dynamic granularity encoding scheduler to generate a sequence of dynamic encoding visual tokens $Z_i$.
To ensure policy diversity and avoid convergence to a local optimum, we implement three key strategies: (1) we iteratively prompt the LLM with the instruction `try another grounding policy different from the history ones' to generate distinct temporal segments; (2) we employ a counterfactual augmentation technique where the inverse of a generated policy (i.e., treating all non-selected segments as the positive result) is included as a valid policy candidate; and (3) we randomly generate several policies based on the video length and vary random seeds during each training sample. This multifaceted approach to policy generation prevents the DPO optimization from collapsing into a limited set of predictions and encourages the model to explore a broader and more robust solution space for segment grounding.

\begin{equation}
\begin{split}
\label{eq:grounding_output}
    output&=VidLLM(input)\\
    &=[[s_1,e_1],...,[s_j,e_j]],
\end{split}
\end{equation}

Generally, compared to the token sequence representing irrelevant segments in fine-grained or relevant segments in coarse-grained, we can assert that a dynamic visual tokens sequence correctly represents all the question-related key video frames in a fine-grained manner and other irrelevant video frames only occupy a small amount of visual tokens through coarse-grained encoding, would be more helpful for the video LLM to understand a multi-event long video, since it provides detailed spatial-temporal visual context precisely where it is needed for detailed understanding and temporal reasoning, while decreases the noise from the global video information by compressing the irrelevant segments.

Therefore, if we input the video question with each sequence of dynamic encoding visual tokens to the video LLM, the correct sequence of visual tokens would help the video LLM to generate an answer closer to the ground truth of the video question, and the cross-entropy loss between them would be smaller. While an incorrect sequence of visual tokens tends to mislead the video LLM, resulting in an incorrect answer response and an increase in loss. 

Inspired by the reinforcement learning from human feedback~\cite{christiano2017deep}(RLHF), we apply direct preference optimization~\cite{rafailov2024direct}(DPO) based on the different cross-entropy scores of the visual token sequence inputs from the grounding policies. This method optimizes our MoD-VLLM framework without explicitly rewarding models and formulates the objective as:
\begin{equation}
\label{equ:dpo}
\small
\begin{split}
    &L_{DPO}(\pi_{\theta};\pi_{ref})=
    \\&-E_{(q,v,p_w,p_l)\sim \mathcal{D}}[\log \sigma(\beta\log\frac{\pi_{\theta}(p_w|q)}{\pi_{ref}(p_w|q)}-\beta\log\frac{\pi_{\theta}(p_l|q)}{\pi_{ref}(p_l|q)})],
\end{split}
\end{equation}
where $p_w$ represents the positive grounding policy leading to the smaller cross-entropy loss, and $p_l$ denotes the negative policy with the larger loss. $\pi_{\theta}$ represents the video LLM to be optimized in this stage, and $\pi_{ref}$ is a reference model initialized with the same video LLM but remains frozen.$\sigma$ is the sigmoid function, and $\beta$ is a controlling parameter. 

As shown in Algorithm~\ref{alg:dynamic-dpo}, our dynamic granularity reinforcement learning alternates between the DPO of grounding policy generation and the SFT of dynamic representation learning. We first employ direct preference optimization to enable the video LLM for more accurate grounding policy generation, and then adopt supervised fine-tuning on the video LLM to adapt the dynamic granularity visual token sequence input.


\begin{algorithm}
\small
\caption{Dynamic Granularity Reinforcement Learning}
\label{alg:dynamic-dpo}
\begin{algorithmic}[1]
\Require Video LLM $VidLLM$, Dynamic Granularity Module $D$, Positive-Negative Grounding Module $G$, dataset $\mathcal{D}=\{(v_i,q_i,y_i)\}^N_{i=1}$, training steps $Step$, gradient accumulate step $s$, numbers of policies per data $n$

\State \textbf{Activate:} $VidLLM$
        
\For{$t=1$ \textbf{to} $Step$}
    \For{$m=1$ \textbf{to} $s$}
        \State initialize $\pi_\theta=VidLLM,\pi_{ref}=VidLLM$
        \State $i \gets ((t-1)s + m - 1) \% N + 1$
        \State Prepare data $(v_i,q_i,y_i)$ from $\mathcal{D}$
        \State Generate policies $p_1,p_2,...p_n = G(\pi_{ref},q_i,v_i)$
        \For{$j=1$ \textbf{to} $n$}
            \State \textbf{Grounding}: $S_j = G(v_i,p_j)$
            \State \textbf{Dynamic granularity encoding}: $v_j = D(S_j)$
            \State \textbf{Forward propagation}: $\widehat{y} = VidLLM(q_i,v_j)$
            \State \textbf{Compute $L_{CEj}=- y \log(\widehat{y}(q,v_j))$}
        \EndFor
        \State $p_w \gets \arg\min_{\{p_j\}} L_{CE}$
        \State $p_l \gets \{p_j, p \neq p_w\}$
        \State \textbf{Optimize $\pi_\theta$ with DPO loss}
    \EndFor
    \State $VidLLM\leftarrow\pi_{\theta}$
    \For{$m=1$ \textbf{to} $s$}
        \State $i \gets ((t-1)s + m - 1) \% N + 1$
        \State Prepare data $(v_i,q_i,y_i)$ from $\mathcal{D}$
        \State Generate single policy $p = G(VidLLM,q_i,v_i)$
        \State \textbf{Grounding}: $S_j = G(v_i,p_j)$
        \State \textbf{Dynamic granularity encoding}: $v_j = D(S_j)$
        \State \textbf{Forward propagation}: $\widehat{y} = VidLLM(q_i,v_j)$
        \State \textbf{Optimize $VidLLM$ with cross-entropy loss}
    \EndFor
\EndFor
\end{algorithmic}
\end{algorithm}

\subsection{MEventBench.} 

\begin{figure}[h]
    \centering
    \includegraphics[width=\linewidth]{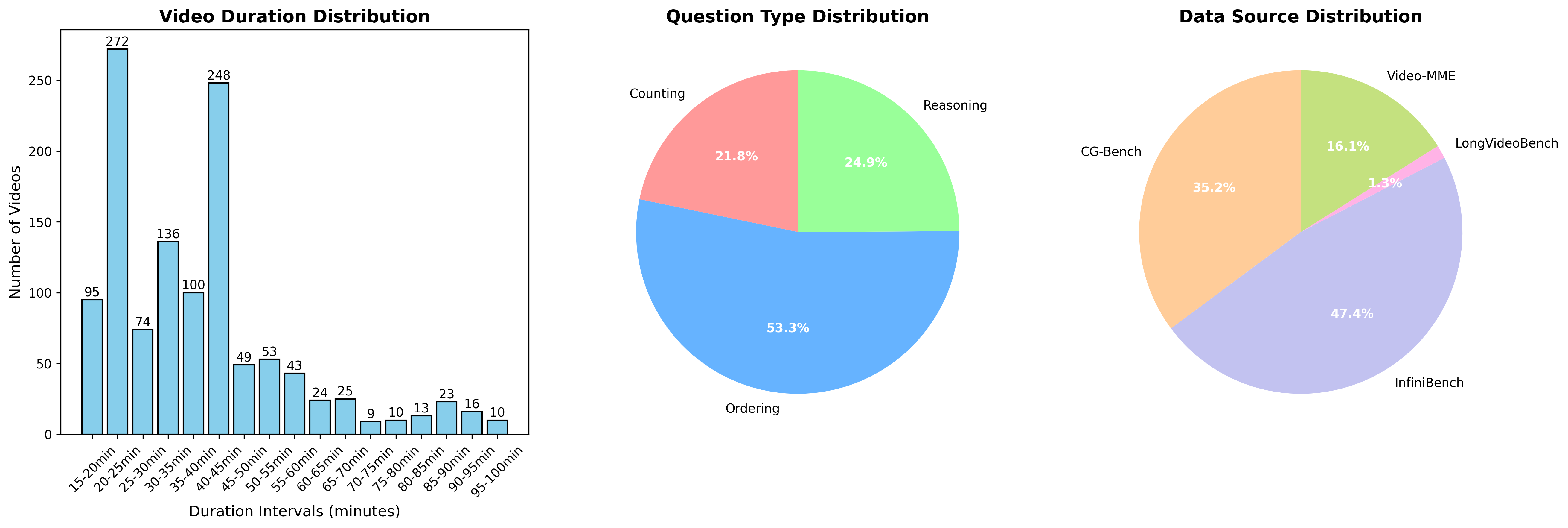}
    \caption{Data distribution of MEventBench.}
    \label{fig:distribution}
\end{figure}

Here, we provide the detailed data distribution of our proposed MEventBench dataset in Figure~\ref{fig:distribution}.
We construct the multi-event long-video dataset as follows. First, we filter existing video benchmarks to retain only videos longer than 10 minutes. From these, we select samples whose original type-annotations include keywords such as “ordering,” “counting,” or “reasoning.” For the remaining data, we search the question texts for indicative phrases like “how many,” “what order,” and “why.” The union of these two subsets yields several thousand candidate video–question pairs. We then manually inspect each candidate to ensure that the question indeed refers to multiple, temporally dispersed segments within the video rather than a single key video segment. The final dataset is assembled from the pairs that satisfy this multi‑segment requirement.
In summary, we classify the type of our selected multi-event data into multi-event counting, ordering, and reasoning.

\textbf{Counting tasks.} These data require models to identify and enumerate multiple instances of specific events or objects distributed sparsely throughout long videos. The key challenge lies in accurately detecting and tallying occurrences that may be temporally dispersed, visually similar, or partially occluded, demanding robust temporal localization and fine-grained visual discrimination.

\textbf{Ordering tasks.} These data focus on temporal ordering by requiring models to arrange multiple events based on their occurrence patterns chronologically. The complexity arises from events with ambiguous temporal boundaries, overlapping timelines, or subtle contextual cues that determine their sequential relationships, necessitating precise temporal understanding.

\textbf{Reasoning tasks.} These data emphasize causal and logical inference by challenging models to identify underlying relationships between multiple events, such as cause-effect chains, prerequisite conditions, or intentional motivations. The difficulty stems from the need to integrate disparate visual evidence across long temporal spans and construct coherent narrative structures from sparse, distributed events.

Together, these three task types form a comprehensive evaluation framework that assesses models' capabilities in quantitative analysis (counting), temporal understanding (ordering), and higher-level cognitive processing (reasoning), covering the essential dimensions of complex multi-event video comprehension.

\end{document}